%% file: ms.tex
\documentclass{article}

\usepackage[final]{tackling_climate_workshop_style}

\usepackage[utf8]{inputenc} 
\usepackage[T1]{fontenc}    
\usepackage{hyperref}       
\usepackage{url}            
\usepackage{booktabs}       
\usepackage{amsfonts}       
\usepackage{nicefrac}       
\usepackage{microtype}      

\usepackage{comment}

\usepackage{natbib} 
    \setcitestyle{numbers}
\usepackage[utf8]{inputenc} 
\usepackage[T1]{fontenc}    

\usepackage{url}            
\usepackage{booktabs}       
\usepackage{amsfonts}       
\usepackage{nicefrac}       
\usepackage{microtype}      
\usepackage{xcolor}         

\usepackage{graphicx}
\usepackage[american]{babel}
\usepackage{nicefrac}       
\usepackage{amsmath}
\usepackage{amsthm}
\usepackage{bm}
\usepackage{blkarray}

\usepackage{thm-restate}
\usepackage{caption}
\usepackage{wrapfig}

\newcommand\bx{\bm{x}}

\newcommand\bz{\bm{z}}
\newcommand\method{CDSD}
\input{math_commands.tex}

\usepackage[nameinlink,capitalize]{cleveref}
\creflabelformat{equation}{#1#2#3}
\Crefname{equation}{Equation}{Equations}
\Crefname{section}{Section}{Sections}

\usepackage{subfigure}
\usepackage[labelfont=bf]{caption}

\usepackage[subtle,mathspacing=normal]{savetrees}

\title{Towards Causal Representations of Climate Model Data}

\author{%
    Julien Boussard$^{1,3 *}$ \quad Chandni Nagda$^{1,4 *}$ \quad Julia Kaltenborn$^{1,5}$ \quad 
    \\
    \textbf{Charlotte Emilie Elektra Lange}$^{1,6}$ \quad \textbf{Philippe Brouillard}$^{1,7}$  
    \\ 
    \textbf{Yaniv Gurwicz}$^{2}$ \quad \textbf{Peer Nowack}$^{8}$ \quad \textbf{David Rolnick}$^{1,5,7}$
    \\
    \\
    $^1$Mila --Quebec AI Institute $^2$Intel Labs $^3$Columbia University \\$^4$University of Illinois at Urbana-Champaign $^5$McGill University $^6$Osnabrück University \\ $^7$Université de Montréal $^8$Karlsruhe Institute of Technology *-equal contribution
}

\begin{document}

\maketitle

\begin{abstract}
Climate models, such as Earth system models (ESMs), are crucial for simulating future climate change based on projected Shared Socioeconomic Pathways (SSP) greenhouse gas emissions scenarios. While ESMs are sophisticated and invaluable, machine learning-based emulators trained on existing simulation data can project additional climate scenarios much faster and are computationally efficient. However, they often lack generalizability and interpretability. This work delves into the potential of causal representation learning, specifically the \emph{Causal Discovery with Single-parent Decoding} (CDSD) method, which could render climate model emulation efficient \textit{and} interpretable. We evaluate CDSD on multiple climate datasets, focusing on emissions, temperature, and precipitation. Our findings shed light on the challenges, limitations, and promise of using CDSD as a stepping stone towards more interpretable and robust climate model emulation.
\end{abstract}


\section{Introduction}

Climate models are indispensable for simulating future climate scenarios based on Shared Socioeconomic Pathways (SSP) emissions. Earth system models (ESMs) are complex models grounded in systems of differential equations that capture a vast array of physical processes. They provide a comprehensive understanding of climate dynamics, but are computationally expensive, as even a single ESM run for one SSP requires about a year to run on a supercomputer \citep{balaji2017cpmip}. 
Recent advancements in data-driven models using machine learning (ML) present an opportunity to emulate climate projections more efficiently \citep{nguyen2023climax, watson2022climatebench}. However, these climate model emulators often act as ``black boxes'', lacking interpretability crucial for climate science.

Causal methods enable the discovery and quantification of causal dependencies in observed data \citep{runge2023causal, rohekar2023from}, and have emerged as a valuable tool to improve our understanding of physical systems across various fields, including Earth Sciences (see \citep{runge2019inferring}). In regards to climate modeling, causal methods can potentially bridge the gap between well defined, but compute-intensive ESMs and efficient, but ``black-box'' ML models. They could be used for A) causal evaluation of climate model emulators, by identifying causal dependencies within their projections and verifying that they correspond to known physical processes; B) climate emulation, by inferring causally-informed high-level representations underlying climate projections 
; and C) causal hypothesis testing and attribution of climate change or extreme events. 
In particular for climate model emulation and the evaluation of those emulators, causal methods have not been used yet. In this work, we aim to investigate the potential of causal methods in the context of climate model emulators.
 
Previous work has leveraged causal methods in various forms to increase our understanding of climate data. The necessary foundation for quantifying causality from time-series, was laid by Granger causality \citep{granger_causality}. It was used to infer causal links between CO$_2$ concentration and global temperature \citep{granger_temp} and was later extended to deduce causal feedbacks in climate models \citep{vannes_feedback}. Building on this, approaches like PCMCI and PCMCI+ \citep{runge2015identifying, runge2018conditional, runge2019detecting, runge2020discovering} were developed to autonomously discover the causal graphs of observed systems, finding applications in climate science \citep{nowack2020causal}. More recent work such as Varimax-PCMCI \citep{tibau2022spatiotemporal} combines PCMCI with dimensionality reduction methods 
to derive causally-informed representations from observable climate data.
However, since these methods do not scale well and become computationally inefficient for large and non-linear climate datasets, \citep{brouillard2023cdsd} introduced a differentiable causal discovery method called \emph{Causal Discovery with Single-parent Decoding} (CDSD). 
CDSD uniquely learns both the latent representation from time-series and the causal graph over these latents simultaneously. Here, we are leveraging CDSD to uncover low-dimensional latent drivers that encapsulate temporal processes in climate model data. 

Our primary objective is to harness the power of causal methods to make ML-based climate emulation more interpretable and trusted by experts. As a first step, we apply temporal causal representation learning, in the form of CDSD, to learn causally informed low-dimensional latents underlying the emissions, temperature, and precipitation time-series from ClimateSet \cite{kaltenborn2023climateset}. 
We show that CDSD is able to infer representations that match known physical processes, demonstrating that CDSD can be used to evaluate and validate climate models. 
We believe that our findings and proposed solutions pave the way for further development of causally informed climate model emulation techniques.

\vspace{-0.5em}

\section{Causal Discovery with Single-Parent Decoding}
\vspace{-0.3em}

CDSD considers a generative model where \(d_x\)-dimensional variables \(\{\bx^t\}^T_{t=1}\) are observed across \(T\) time steps. These observations, \(\bx^t\), are influenced by $d_z$-dimensional latent variables \(\bz^t\). For instance, \(\bx^t\) could represent climate measurements, while \(\bz^t\) might represent unknown regional climate trends.

The model considers a stationary time series of order \(\tau\) over these latent variables. Binary matrices \(\left\{G^k\right\}_{k=1}^\tau\) represent the causal relationships between latents at different time steps. Specifically, an element \(G^{k}_{ij} = 1\) indicates that \(z^{t-k}_j\) is a causal parent of \(z^{t}_i\), capturing the lagged causal relations between the time-steps \(t-k\) and \(t\). The adjacency matrix \(F\) delineates the causal connections between the latent variables \(\bz\) and the observed variables \(\bx\). Each observed variable \(x_i\) has at most one latent parent, adhering to the \emph{single-parent decoding} structure. A high-level description of this model is provided here, with comprehensive details presented in \cref{sec:add_inference}.

\begin{figure}[h]
    \centering
    \includegraphics[width=\textwidth]{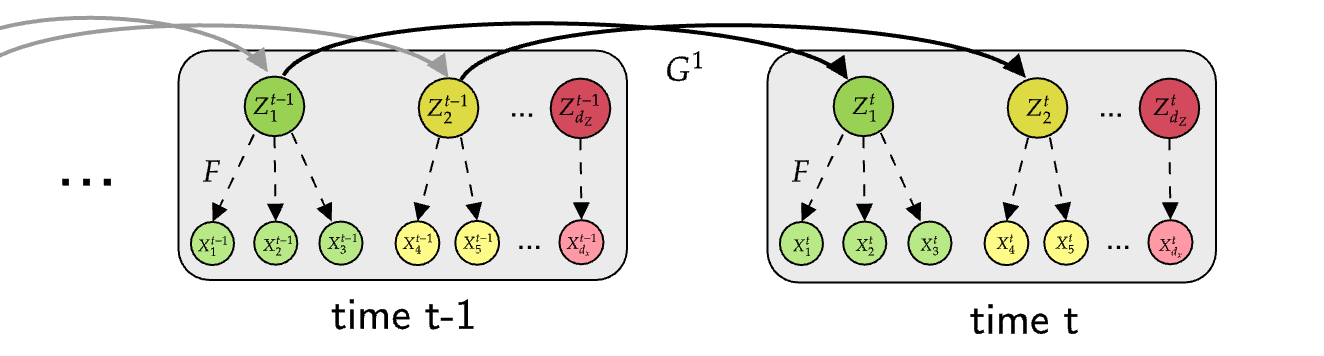}
    \caption{\label{fig:generative_model} \textbf{Generative model.} Variables \(\bz\) are latent, and \(\bx\) are observable. \(G^k\) represents latent connections across different time lags, with the diagram only containing connections up to \(G^1\). \(F\) connects latents to observables. Connections are illustrated only up to \(G^1\), but \method{} leverages connections of higher order. \textit{Figure reprinted with permission from \cite{brouillard2023cdsd}.}}
\end{figure}
\vspace{0.8em}
At any given time step \(t\), the latents are assumed to be independent given their past, and each conditional is parameterized by a non-linear function $g_j$. $h$ is chosen to be a Gaussian density function.
\vspace{-0.8em}
\begin{equation} \label{eq:facto_transition}
   p(\bz^t \mid \bz^{t-1}, \dots, \bz^{t-\tau}) := \prod_{j=1}^{d_z} p(z^t_{j} \mid \bz^{t-1}, \dots, \bz^{t-\tau}); 
\end{equation}
\begin{equation} \label{eq:conditional_transition}
    p(z^t_{j} \mid \bz^{<t}) := \\ h(z^t_{j}; \,\,g_j([G^1_{j :} \odot \bz^{t-1}, \dots, G^\tau_{j :} \odot \bz^{t - \tau}])\,),
\end{equation}

The observable variables \(x_j^t\) are assumed to be conditionally independent where $f_j: \mathbb{R} \rightarrow \mathbb{R}$, and $\bm{\sigma}^2 \in \mathbb{R}^{d_x}_{> 0}$ are decoding functions:
\begin{equation} \label{eq:conditional_observation}
    p(x^t_j \mid \bz^t_{pa^F_j}) := \mathcal{N}(x^t_j; f_j(\bz^t_{pa^F_j}), \sigma^2_j),
\end{equation}

The model's complete density is:
\begin{equation}
    p(\bx^{\leq T}, \bz^{\leq T}) := \prod_{t=1}^T p(\bz^t \mid \bz^{<t}) p(\bx^t \mid \bz^t).
\end{equation}

Maximizing $p(\bx^{\leq T}) = \int p(\bx^{\leq T}, \bz^{\leq T}) \, d\bz^{\leq T}$ unfortunately involves an intractable integral, hence the model is fit by maximizing an evidence lower bound (ELBO) \citep{kingma2013auto, girin2020dynamical} for \(p(\bx^{\leq T})\). The variational approximation of the posterior \(p(\bz^{\leq T} \mid \bx^{\leq T})\) is \(q(\bz^{\leq T} \mid \bx^{\leq T})\).
\begin{equation} \label{eq:conditional_posterior}
    q(\bz^{\leq T} \mid \bx^{\leq T}) := \prod_{t=1}^T q(\bz^t \mid \bx^t); \,\,\,\,\,    q(\bz^t \mid \bx^t) := \mathcal{N}(\bz^t; \tilde{\bm{f}}(\bx^t), \text{diag}(\tilde{\bm{\sigma}}^2)),
\end{equation}
\begin{equation} \label{eq:elbo_climate}
    \log p(\bx^{\leq T}) \geq \sum_{t=1}^T \Bigl[ \mathbb{E}_{\bz^t \sim q(\bz^t \mid \bx^t)}\left[\log p(\bx^t \mid \bz^t)\right] - \\
    \mathbb{E}_{\bz^{<t} \sim q(\bz^{<t} \mid \bx^{<t})} \text{KL}\left[q(\bz^t \mid \bx^t) \,||\, p(\bz^t \mid \bz^{< t})\right] \Bigr].
\end{equation}

The graph between the latent \(\bz\) and the observable \(\bx\) is parameterized using a weighted adjacency matrix \(W\). To enforce the single-parent decoding, \(W\) is constrained to be non-negative with orthonormal columns. Neural networks are optionally used to parameterize encoding and decoding functions $g_j$, $f_j$, $\tilde{\bm{f}}$. The graphs \(G^k\) are sampled from \(G^k_{ij} \sim Bernoulli(\sigma(\Gamma^k_{ij}))\), with \(\Gamma^k\) being learnable parameters. The objective is optimized using stochastic gradient descent, leveraging the Straight-Through Gumbel estimator ~\citep{maddison2016concrete, jang2016categorical} and the reparameterization trick \citep{kingma2013auto}. See \cref{sec:add_inference}
for more details. 

\vspace{-0.5em}

\section{Results on Climate Data}\label{sec:exp-climate}
\vspace{-0.3em}

\begin{figure}[t]
    \centering
    \includegraphics[width=\textwidth]{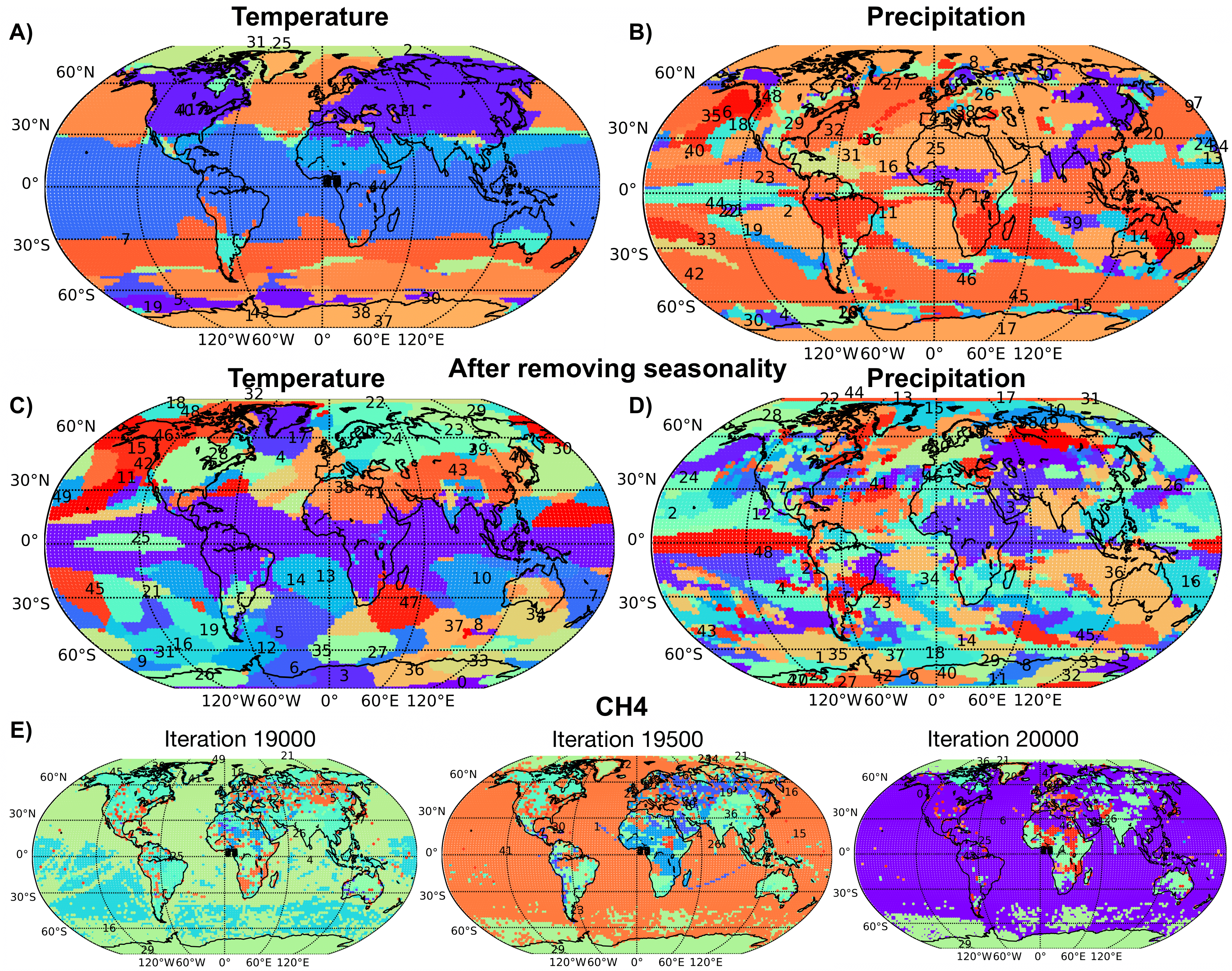}
    \caption{\label{fig:results} Segmentation of the Earth's surface according to the mapping between observations and latents learnt by CDSD for \textbf{(A)} CMIP6 temperature data of SSP2-4.5; \textbf{(B)} CMIP6 precipitation data of SSP2-4.5; \textbf{(C)} CMIP6 temperature and precipitation data of SSP2-4.5 after subtracting seasonal trends; and \textbf{(D)} CH$_4$ Input4MIPs data, at 3 different iterations after the loss plateaus. Each location is colored according to its parent in the causal representation.}
\end{figure}

Our experiments use emission, temperature, and precipitation data from ClimateSet \cite{kaltenborn2023climateset}, which extends ClimateBench \cite{watson2022climatebench}, and is sourced from the Coupled Model Intercomparison Project Phase 6 (CMIP6) \citep{cmip6_overview} and the Input Datasets for Model Intercomparison Projects (Input4MIPs) \citep{durack2017input4mips}. From CMIP6, we selected the temperature and precipitation outputs of the Nor-ESM2 model, and from Input4MIPs the CO$_2$, CH$_4$, SO$_2$, and black carbon (BC) emission input data. For all input and output data the SSP2-4.5 scenario was chosen.
For all variables, we consider the time-series spanning 2015 -- 2100, with monthly frequency. Each time step corresponds to a $144 \times 96$ spatial global map (250~km resolution). For the experiments presented in \cref{fig:results}, we ran CDSD on this data using a latent dimension $d_z=50$, and inputs of time-length $\tau=5$. In other words, we want to discover up to 50 latent variables, with causal links emerging over 5 months. Additional parameters and experimental setup are detailed in \cref{app_params}.

\cref{fig:results}(A) shows the clustering induced by CDSD for temperature across all grid point locations on the globe. Clusters represent grid points across the globe that have a common parent in the latent causal graph. These clusters can be interpreted as regions with similar climatic properties.
Causal graphs over the latents represent the potential causal links between these spatial points across different time lags. Most clusters are found outside the tropics. This grouping can be understood from the larger seasonal variation in temperatures at higher latitudes, again combined with also larger high-latitude warming under greenhouse gas forcing (e.g. polar amplification \citep{polaramp1, polaramp2}). There is a clear land-ocean separation in the northern hemisphere. The causal clusters and linkages uncovered by CDSD on temperature correspond to the different climatic zones on Earth and captures their seasonal trends. According to our analysis, CDSD does not find meaningful causal connections between the recovered clusters and captures only seasonal variations but no forced trends. One possible explanation is that seasonal variations are magnitudes larger than forced trends, impeding the recovery of the trends relevant to climate modeling questions. 


Similarly, \cref{fig:results}(B) shows the clustering induced by CDSD when ran on precipitation data. Clusters appear spread out and less compact than those obtained with temperature data.
This can be attributed to the nature of precipitation data, which is more patterned and regional (relatively wet and dry regions occur both at high and low latitudes). The clusters also highlight several semi-permanent large-scale high and low pressure systems, reflecting regions of low and high total rainfall respectively. The ``dynamic'' movement of air masses that are much more influential on precipitation than on temperature explains the elongated shape of many clusters. As weather systems move across ocean and land masses, there is not as clear a land-ocean separation as for temperature. Dry areas such as the Sahara desert, and rainforest areas, such as Amazon and Congo Basin, with high levels of average precipitation are visible. Large-scale atmospheric circulation phenomena governing precipitation, such as convergence zones become visible, as well as clusters following tropical and extra-tropical cyclone patterns and their storm tracks. Clusters seem to broadly recover the Intertropical Convergence Zone (ITCZ) seasonal differences, as ITCZ is close to the equator during winter and moves above India and through China during the Northern Hemisphere's summer months.

To capture physical phenomena independent of seasonal variations, we perform ``seasonality removal'', and normalize the data by subtracting the monthly mean and dividing by the monthly standard deviation. The resulting clusters for temperature data, shown in \cref{fig:results}(C) appear significantly different. Climatic zones - especially within tropical, subtropical and temperate zones - are now recovered by CDSD for temperature data. While there is still a dominant cluster along the equator, CDSD now finds a mid-Pacific cluster associated with the El Niño Southern Oscillation (ENSO), cluster 25, which corresponds to cluster 44 in (B).
The zonal tropospheric circulation - see clusters surrounding the Antarctica in (A) and (B) - is more clearly patterned when removing seasonality. As the location and strength of those pressure cells is varying distinctly with seasons, their corresponding clusters might become more pronounced and cover more space when removing seasonality.

When seasonality is not removed from the precipitation data, the observed patterns closely align with sea level pressure maps, capturing broader climatic trends. However, by removing seasonality, the CDSD model reveals more localized climate phenomena. We postulate that in this more difficult regime, the model cannot rely on large seasonal rainfall differences to form clusters, and thus captures more granular detail. For instance, Antarctica is segmented into a greater number of clusters when seasonality is removed (\cref{fig:results} (D)). CDSD may be picking up on transient storm systems influenced by the phase behavior of the Southern Annular Mode (SAM). SAM is known for causing a poleward shift of storm tracks, as well as increased precipitation in the southern parts of Australia and South America. Furthermore, rainforest clusters appear more detailed, with the separation between tropical and subtropical parts of the rainforests becoming more apparent. 
Southeast Alaska, the northern-most temperate zone of the Northern Hemisphere also appears separately as part of the purple cluster (24). Cluster 30, the light-green cluster over western-central Europe, likely corresponds to the storm tracks of eastwards-moving wet weather systems from the North Atlantic into Europe. Other phenomena are less clear. For example, clusters over Antarctica are very sensitive to parameterization: In some runs Antarctica's orography (``landscape'') seems to influence local precipitation patterns strongly, while in others those patterns are less clear. Such phenomena need to be evaluated in further experiments. 

\cref{fig:results}(E) shows the clustering induced by CDSD ran on CH$_4$ emissions data, at different iterations before convergence. The model converges after 20000 iterations, according to the convergence criteria (detailed in \cref{app_params}) of the optimized loss ( \cref{eq:full_objective} in \cref{sec:add_inference}). At each iteration shown in the figure, the loss is very close to the convergence loss, but the induced clustering is very different, showing that multiple representations correspond to similar objective values. CDSD is not able to robustly capture ships' CH$_4$ emissions, sometimes represented by clusters forming lines between different ports. It is not a dominant part of CH$_4$ emissions, but clearly shows that CDSD does not converge. 
We can see very different results in regions such as South-East Asia, America or oceans, with very different number of clusters being found at each iteration. It seems harder to find a stable causal representation underlying anthropogenic emission data than for physical climate variable data.
\vspace{-0.5em}

\section{Discussion}
\vspace{-0.3em}

\method{} is able to represent meaningful physical processes of temperature and precipitation measurements on the seasonal-to-century scale. We demonstrate results that coincide with known phenomena, such as regional temperature trends, ENSO, tropical and extra-tropical cyclone routes. We have highlighted the need for removing large seasonal variations that otherwise dominate forced trends. This approach could potentially remove the imprint of the seasonal cycle. To avoid this potential failure, one could standardize data with respect to the standard deviation and variable-dependent average over the complete time-series. Future work will explore modified approaches to distinguish forced and seasonal trends. 

\method{} fails to stably represent emissions data. This makes sense, since forcing agent emissions are dominantly driven by anthropogenic effects and not natural physical processes. Human policy decisions and economic activities do not adhere to predictable temporal causal relationships and are thus not recoverable by CDSD. Furthermore, the lifetimes of different gases and aerosols range from weeks to hundreds of years, and impact the climate at various spatio-temporal resolutions. For example, carbon dioxide can persist in the atmosphere for thousands of years, leading to long-term cumulative warming effects. On the other hand, methane has a relatively short atmospheric lifetime of less than 12 years, but traps more heat \cite{ipcc2022}. These discrepancies make for a particularly challenging task, as \method{} expects evenly sampled inputs with consistent time resolutions and expects causal temporal relationships to manifest within the relatively small time length $\tau$. However, the number of parameters in the model scales linearly with $\tau$; hence, increasing it beyond several years is computationally expensive and will make convergence difficult. 
Currently, we run \method{} using $\tau = 5$. For future work, we plan to use Transformers or Long Short-Term Memory to parameterize the transition functions $g_j$ to handle larger values of $\tau$. Also, as emissions are a cause of changes in observed climate variables, it might be possible to hard-code this causal relation and condition the learnt representation on the emissions. Such changes might allow \method{} to represent forced trends and improve climate projection.



\vspace{-0.5em}

\section{Conclusion}
\vspace{-0.3em}

Here, we demonstrated a first successful application of CDSD to investigate causal links in climate model data, and highlight future challenges in applying CDSD to emission data. For temperature and precipitation, the learned representations could be used to compare and evaluate different climate models and/or observations. Using CDSD in ML-based climate model emulators remains a challenge, albeit with the potential to render those emulators, for the first time, interpretable. Future work will consider crucial next steps (e.g., timescales of interest) towards such efficient and interpretable emulators, which are needed for the climate modeling community and ultimately to help inform policymakers.




\pagebreak
\bibliography{ms}

\clearpage
\appendix

\section{Inference with \method{}: Objective and Optimization}\label{sec:add_inference}

In this section, we present how inference and optimization is carried out when using \method{} \cite{brouillard2023cdsd}.

\textbf{Continuous optimization.~~}The graphs $G^k$ are learnt via continuous optimization. They are sampled from distributions parameterized by $\Gamma^k \in \mathbb{R}^{d_z \times d_z}$ that are learnable parameters. Specifically, $G^k_{ij} \sim Bernoulli(\sigma(\Gamma^k_{ij}))$, where $\sigma(\cdot)$ is the sigmoid function. 
This results in the following constrained optimization problem, with $\bm{\phi}$ denoting the parameters of all neural networks ($r_j$, $g_j$, $\tilde{\bm{f}}$) and the learnable variance terms at Equations~\ref{eq:conditional_observation} and \ref{eq:conditional_posterior}:

\begin{equation}\label{eq:const-opt-prob}
\begin{aligned}
    & \max_{W, \Gamma, \bm{\phi}} \mathbb{E}_{G \sim \sigma(\Gamma)} \bigl[\mathbb{E}_{\bx}\left[\mathcal{L}_{\bx}(W, \Gamma, \bm{\phi}) \right] \bigr] - \lambda_s ||\sigma(\Gamma)||_1 \\
    & \text{s.t.} \ \ W \ \text{is orthogonal and non-negative} ,
\end{aligned}
\end{equation}

$\mathcal{L}_{\bx}$ is the ELBO corresponding to the right-hand side term in \cref{eq:elbo_climate} and $\lambda_s > 0$ a coefficient for the regularisation of the graph sparsity. The non-negativity of $W$ is enforced using the projected gradient on $\mathbb{R}_{\geq 0}$, and its orthogonality enforced using the following constraint:

\begin{equation*}
    h(W) := W^T W - I_{d_z} \ .
\end{equation*}

Thsi results in the final constrained optimization problem, relaxed using the \textit{augmented Lagrangian method} (ALM):
\begin{equation} 
\label{eq:full_objective}
    \max_{W, \Gamma, \bm{\phi}} \mathbb{E}_{G \sim \sigma(\Gamma)} \bigl[\mathbb{E}_{\bx}[\mathcal{L}_{\bx}(W, \Gamma, \bm{\phi})]\bigr] \\ - \lambda_s ||\sigma(\Gamma)||_1 -  \text{Tr}\left(\lambda_W^T h(W)\right) - \frac{\mu_W}{2} || h(W) ||^2_2 ,
\end{equation}
where $\lambda_W \in \mathbb{R}^{d_z \times d_z}$ and $\mu_W \in \mathbb{R}_{> 0}$ are the coefficients of the ALM. 

This objective is optimized using stochastic gradient descent. The gradients w.r.t. the parameters $\Gamma$ are estimated using the Straight-Through Gumbel estimator~\citep{maddison2016concrete, jang2016categorical}. The ELBO is optimized following the classical VAE models~\citep{kingma2013auto}, by using the reparametrization trick and a closed-form expression for the KL divergence term since both $q(\bz^t \mid \bx^t)$ and $p(\bz^t \mid \bz^{< t})$ are multivariate Gaussians. The graphs $G$ and the matrix $W$ are thus learnt end-to-end. 

\section{Detailed Parameters and Experimental Setup}\label{app_params}

We train our models on our internal cluster and use a single Nvidia-RTX8000 with 32GB of RAM for each run. 

Reusing the default parameters of \method{} detailed in \citep{brouillard2023cdsd}, we use leaky-ReLU as activation functions for all neural networks. For the neural networks $g_j$ fitting the non-linear dynamic, we used Multi-layer perceptrons (MLPs) with 2 hidden layers and 8 hidden units. For the neural network $r_j$ fitting the non-linear encoding, we use a single neural network that receives as input the masked $W\bz^t$ and an embedding (dimension 10) of the index $s(j)$ is concatenated to the input. This neural network has 2 hidden layers and 32 hidden units. Furthermore, we use the optimizer RMSProp~\citep{hinton2012neural}. 

As we encountered problems with convergence when running experiments on emissions data, we performed a hyperparameter search for learning rate and batch size. For all experiments, we tried learning rates among $\{1e-2, 1e-3, 1e-4\}$, and batch sizes among $\{64, 128, 256, 512\}$. We also tried learning rate decay, without better success. For experiments reported in \cref{fig:results}, we used a learning rate of $1e-3$ and batch size of $64$. For all runs, the data is divided with a split ratio of $0.9$ for the training set and $0.1$ for the validation set. Models are trained until convergence, determined once the validation loss has not improved in $1000$ iterations. As recommended by \citep{brouillard2023cdsd}, we tried multiple values for the regularization coefficient enforcing graph sparsity of CDSD $\{10, 1, 10^{-1}, 10^{-2}, 10^{-3}, 10^{-4}, 10^{-5}\}$.

\section{Additional Experiments}\label{app_experiments}

To validate our hypothesis and get a better understanding of what needs to be done to use causal representation learning models for climate emulation, we conducted multiple experiments using different inputs, data preprocessing and parameters. We report a list of experiments, along with the insight we gained from them. All experiments were conducted with the experimental setting described in \cref{app_params}. 

We tried using latents of dimension $50, 100$ and $200$, but the results did not differ a lot among each other, although training slowed down with increasing latent dimension. The number of latents was particularly relevant when running \method{} on precipitation data, as we expected that the distinct regions within shared precipitation clusters would eventually separate. However, this was not the case. For future work, we suggest to implement a constraint enforcing the connectivity of the clusters in order to represent causal connections between specific regions of the globe more explicitely. 

As mentioned in the main text, we ran \method{} on the CO$_2$, SO$_2$, and black carbon (BC) emission input data from Input4MIPs, on SSP2-4.5. SO$_2$ and BC, two aerosols, have a lifetime of less than $1$ month whereas CO$_2$ is cumulative in nature and lasts more than $300$ years in the atmosphere (given a working carboncycle - otherwise it technically stays forever). These processes, along with CH$_4$ that has a lifetime of approximately $10$ years, interact on very different resolutions. We plan to run \method{} on carbon monoxide (CO) from Input4MIPs, as this gas has a lifetime of $2--3$ months, corresponding to the resolution of the input data. 

To check if CH$_4$ and CO$_2$ could be represented when using lower time-resolution, we aggregated the monthly data to create annual data, and used $\tau = 5$ as well as $\tau=12$ to capture longer-term causal connections. However, this did not solve the convergence issue. It is possible that the number of data points is now too low, as, after aggregating the data, the number of training points is reduced by $12$. We also tried to remove seasonality, by standardizing the data of each of the $12$ months over different years, hoping to capture non-seasonal causal links. For all gases, we tried inputting time-series of multiple resolution, $1$, $3$, $6$ and $12$ months together in order to learn representations that are invariant to time-resolution. For all these experiments, the model did not converge to a single causal representation. 

An additional difficulty might arise from the high spatial resolution of Input4MIPs data. Areas of high natural CH$_4$ emissions (e.g. wetlands) can be positioned naturally next to regions of overall low emissions or high anthropogenic emissions (e.g. cities, areas of fracking etc.). Therefore, spatial homogeneity is not to be expected and may lead to unstable behaviour. Spatially aggregating the data to get coarser resolution and spatial homogeneity may lead to more stable behavior, although we might lose some information (such as anthropogenic vs. natural emissions). We plan to run \method{} using different input spatial resolution, and study the behavior of \method{} further.


\end{document}

%% file: math_commands.tex
\usepackage{amsmath,amsfonts,bm}

\def\1{\bm{1}}

\DeclareMathAlphabet{\mathsfit}{\encodingdefault}{\sfdefault}{m}{sl}
\SetMathAlphabet{\mathsfit}{bold}{\encodingdefault}{\sfdefault}{bx}{n}

\definecolor{mydarkblue}{rgb}{0,0.08,0.45}